\documentclass[journal]{IEEEtran}

\ifCLASSINFOpdf
\else
   \usepackage[dvips]{graphicx}
\fi
\usepackage{url}
\usepackage{graphicx}
\usepackage[american]{babel}

\usepackage{cite}
\usepackage{mathtools}
\usepackage{amsmath, amssymb, amsthm, bm}
\usepackage{booktabs}
\usepackage{textcomp}
\usepackage[colorlinks = true,
            linkcolor = blue,
            urlcolor  = blue,
            citecolor = blue,
            anchorcolor = blue]{hyperref}
\usepackage{wrapfig}
\graphicspath{ {./figures/} }

\usepackage[ruled,vlined]{algorithm2e}

\newcommand{\dd}{\mathop{}\!\mathrm{d}}

\newtheorem{proposition}{Proposition}

\newcommand{\cu}[1]{
	\ifcat\noexpand#1\relax
	\boldsymbol{#1}
	\else
	\mathbf{#1}
	\fi
}

\begin{document}

\title{Temporal Gaussian Process Regression in Logarithmic Time}

\author{Adrien Corenflos,~\IEEEmembership{Student Member,~IEEE},  Zheng Zhao,~\IEEEmembership{Student Member,~IEEE}, and Simo S\"{a}rkk\"{a},~\IEEEmembership{Senior Member,~IEEE}
\thanks{This work was supported Academy of Finland (projects 321900 and 321891) and Aalto ELEC doctoral school.}
\thanks{Adrien Corenflos (\href{mailto:adrien.corenflos@aalto.fi}{adrien.corenflos@aalto.fi}), Zheng Zhao, and Simo S\"{a}rkk\"{a} were with Department of Electrical Engineering and Automation, Aalto University, Finland. }
}

\markboth{Journal of \LaTeX\ Class Files, Vol. 14, No. 8, August 2015}
{Shell \MakeLowercase{\textit{et al.}}: Bare Demo of IEEEtran.cls for IEEE Journals}
\maketitle

\begin{abstract}
The aim of this article is to present a novel parallelization method for temporal Gaussian process (GP) regression problems. The method allows for solving GP regression problems in logarithmic $O(\log N)$ time, where $N$ is the number of time steps. Our approach uses the state-space representation of GPs which in its original form allows for linear $O(N)$ time GP regression by leveraging the Kalman filtering and smoothing methods. By using a recently proposed parallelization method for Bayesian filters and smoothers, we are able to reduce the linear computational complexity of the temporal GP regression problems into logarithmic span complexity. This ensures logarithmic time complexity when run on parallel hardware such as a graphics processing unit (GPU). We experimentally demonstrate the computational benefits on simulated and real datasets via our open-source implementation leveraging the GPflow framework.
\end{abstract}


\IEEEpeerreviewmaketitle

\section{Introduction}
\label{sec:intro}
\begin{figure}[th]
\vspace{-0.1in}
\centering
\includegraphics[width=\linewidth]{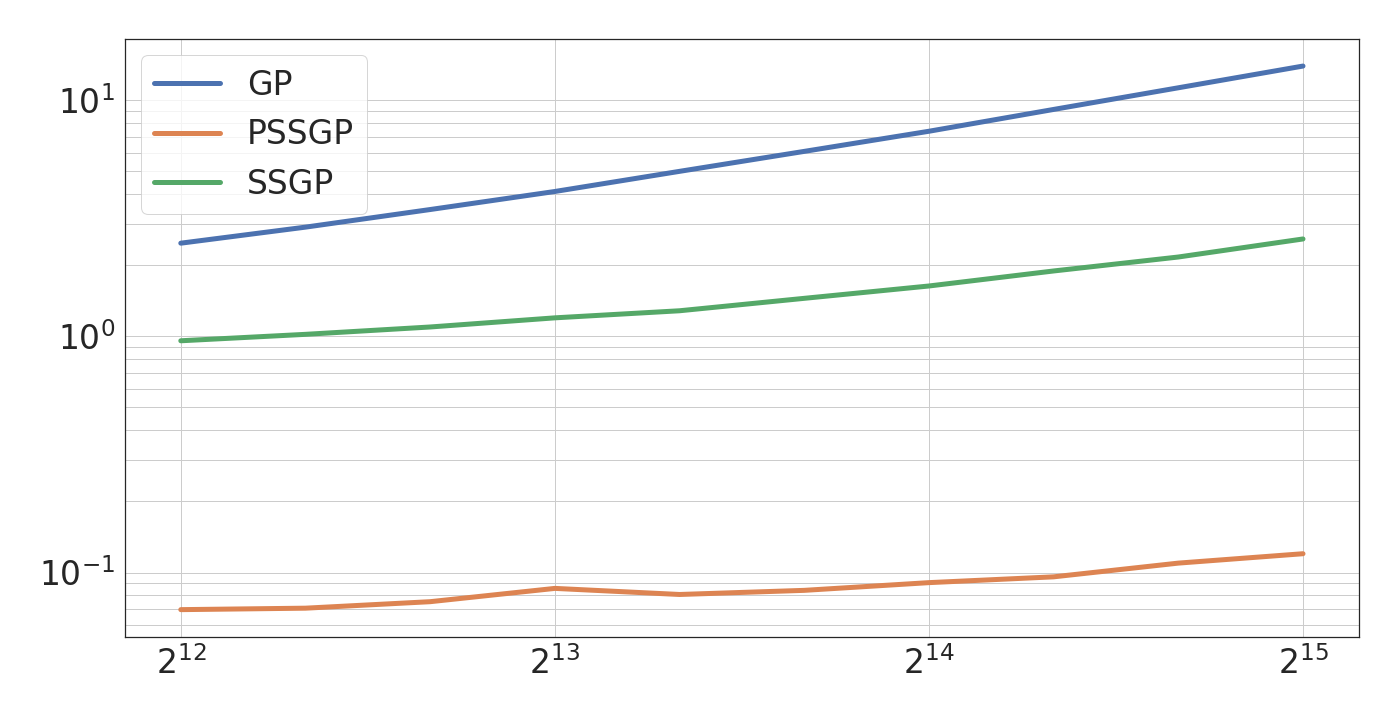}
\caption{Average run time in seconds of predicting $M=10\,000$ test points for noisy sinusoidal data with RBF covariance function for standard GP, SSGP, and PSSGP (ours). $x$-axis is the number of training data points $N$. We can see that our method outperforms GP and SSGP by a factor $>10$.}
\label{fig:concept}
\vspace{-0.1in}
\end{figure}

Gaussian processes (GPs) are a family of useful function-space priors used to solve regression and classification problems arising in machine learning \cite{Rasmussen+Williams:2006}. In their naive form their complexity scales as $O(N^3)$, where $N$ is the number of training data points, which is problematic for large datasets. One approach to alleviating this is to use state-space methods \cite{Hartikainen+Sarkka:2010,Sarkka+Solin+Hartikainen:2013} which reduce the GP regression problem into a Kalman filtering and smoothing problem with linear time complexity $O(N)$ in the number of data points. This linear complexity is optimal on single-threaded computational architectures, as processing data needs to be done sequentially. However it is suboptimal on hardware where parallelization is possible, such as multi-core central processing units (CPUs) or, more importantly, on massively threaded architectures such as graphics processing units (GPU). The aim of this letter is therefore to develop parallel state-space GPs (PSSGPs) methods which reduce the computational complexity (in the sense of parallel span complexity) of state-space GPs to logarithmic $O(\log N)$ (see Fig.~\ref{fig:concept}). To do so we leverage the parallel Bayesian filtering and smoothing methodology of \cite{Sarkka+Garcia:2021}.

Over the recent years, several other approaches to parallelization of GPs have been proposed. For instance in \cite{Liu:2018,Zhang:2019} the authors consider mini-batching the dataset to form mixtures of local GP experts, incurring a cubic cost in the size of the batches. More closely related to this letter are the works in \cite{Grigorievskiy:2017,Lindgren+Rue+Lindstrom:2011} which proposed to leverage the sparse Markovian structure of Markovian and state-space GPs (SSGPs) via the use of parallel matrix computations, thereby reaching $O(\log N)$ span complexity in the dataset size in some special cases. However, the methods outlined in \cite{Grigorievskiy:2017,Lindgren+Rue+Lindstrom:2011} effectively require computations with large (albeit sparse) matrices and the logarithmic span complexity is hard to guarantee for all subproblems \cite{Grigorievskiy:2017}. 
Orthogonally to these parallelization efforts, different approximation methods were introduced in order to reduce the computational complexity of GPs. These include for example inducing points, spectral sampling, or basis function methods (see e.g., \cite{Rasmussen+Williams:2006,Quinonero-Candela+Rasmussen:2005,Lazaro-Gredilla+Quinonero-Candela+Rasmussen+Figueiras-Vidal:2010,Hensman:2018,Solin:2020}). 

The contribution of our paper is three-fold:
\begin{enumerate}
    \item We combine the state-space formulation of GPs with parallel Kalman filters and smoothers \cite{Sarkka+Garcia:2021}.
    \item We extend the parallel formulation to missing measurements needed in prediction with state-space GPs.
    \item We experimentally demonstrate the computational gains of our proposed methods on simulated and real datasets\footnote{We implemented the method as an open-source extensible library. The code can be found at \href{https://github.com/EEA-sensors/parallel-gps}{https://github.com/EEA-sensors/parallel-gps}.}.
\end{enumerate}

\section{Gaussian processes in state-space form}

In this section, we quickly recall results about the state-space formulation of Gaussian processes before presenting their temporal parallelization formulation.  Given covariance function $C(t,t')$ and observations $\lbrace y_k \colon k=1,\ldots,N\rbrace$, a temporal GP regression problem of the form
\begin{equation}\label{eq:gp-def}
\begin{split}
  f(t) &\sim \mathrm{GP}(0, C(t,t')), \\
  y_{k} &= f(t_k) + e_k, \qquad e_k \sim \mathcal{N}(0, \sigma_k^2),
\end{split} 
\end{equation}
can be converted into a smoothing problem for a {$n_{x}$-dimensional} continuous-discrete state-space model
\begin{equation}
\begin{split}
\frac{\dd \cu{x}(t)}{\dd t} = \cu{G}\, \cu{x}(t) + \cu{L}\, \cu{w}(t), \quad
y_k = \cu{H} \, \cu{x}(t_k) + e_k,
\end{split}
\label{eq:ssmodel}
\end{equation}
where $\cu{x}(t)$ is the state, $y_k$ is the measurement, $\cu{w}(t)$ is a white noise process with a spectral density $\cu{Q}$, and $e_k$ is the Gaussian measurement noise~\cite{Hartikainen+Sarkka:2010,Sarkka+Solin+Hartikainen:2013}. The dimensionality $n_x$ of the state as well as the matrices $\cu{G}$, $\cu{L}$, $\cu{H}$, and $\cu{Q}$ in the model depend on (and define) the covariance function at hand. 

In the state-space formulation~\eqref{eq:ssmodel}, the Gaussian process in \eqref{eq:gp-def} has the representation $f(t) = \cu{H} \, \cu{x}(t)$. In the case of Mat\'{e}rn covariance functions this representation happens to be exact and available in closed form \cite{Hartikainen+Sarkka:2010}, while more general stationary covariance functions can be approximated up to an arbitrary precision by using Taylor series or P\'{a}de approximants \cite{Hartikainen+Sarkka:2010,Sarkka+Piche:2014,Solin+Sarkka:2014a,Solin+Sarkka:2014b,Karvonen+Sarkka:2016,Tronarp+Karvonen+Sarkka:2018} in spectral domain. 

The continuous-time state-space model \eqref{eq:ssmodel} can be discretized into an equvalent discrete-time linear Gaussian state space model (LGSSM, e.g., \cite{Sarkka+Solin:2019}) of the form
\begin{equation}
\cu{x}_k = \cu{F}_{k-1} \, \cu{x}_{k-1} + \cu{q}_{k-1}\quad
y_k = \cu{H} \, \cu{x}_k + e_k,
\label{eq:dmodel}
\end{equation}
where $\cu{q}_{k-1} \sim \mathcal{N}(\cu{0}, \cu{Q}_{k-1})$. The GP regression problem can now be solved by applying Kalman filter and smoother algorithms on model \eqref{eq:dmodel} in $O(N)$ time \cite{Hartikainen+Sarkka:2010}. Furthermore, parameters of this model, such as the hyperparameters $\cu{\theta}$ of the covariance function $C(t,t';\cu{\theta})$ can be estimated with standard state-space methods (see, \cite{Sarkka:2013}).

The aim of this paper is to point out that these sequential Kalman filters and smoothers as well as the parameter estimation methods can be replaced by their parallel versions~\cite{Sarkka+Garcia:2021} which reduces the computational complexity to $O(\log N)$. Additionally, parameter estimation using methods such as gradient-based optimization \cite{Nocedal:1999} or Hamiltonian Monte Carlo (HMC) \cite{Neal:2011} requires computation of gradients of the log-likelihood function. However, the same libraries that allow for efficient parallel implementation of the Kalman filters and smoothers such as TensorFlow \cite{tensorflow2015-whitepaper} also support automatic differentiation which can be used to compute the gradients without manual effort and within the parallel computations. 

\section{Handling Missing Observations in Parallel Kalman Filter}
\label{sec:par-kalman}
The formulation in terms of a continuous state space model presented in the previous section reduces the computational complexity of inference in compatible GPs to $O(N)$ instead of the cubic cost induced by naive GP methods, or to a lesser extent by their sparse approximations \cite{Rasmussen+Williams:2006}. 
In \cite{Sarkka+Garcia:2021}, the authors introduce an equivalent formulation of Kalman filters and smoothers in terms of an associative operator which enables them to leverage distributed implementations of scan (prefix-sum) algorithms such as \cite{Blelloch:1989} and \cite{Blelloch:1990}, reducing the time complexity of Kalman filtering and smoothing down to $O(\log N)$.
However, this formulation does not consider the possibility of missing observations. This prevents its application for inference in state-space GP models where test points are treated as missing data~\cite{Sarkka+Solin+Hartikainen:2013}.
    
The method in \cite{Sarkka+Garcia:2021} consists in writing the filtering step in terms of an associative operator of a sequence of five elements $(\cu{A}_k, \cu{b}_k, \cu{C}_k, \cu{\eta}_k, \cu{J}_k)$, which are first initialized in parallel and then combined using parallel associative scan \cite{Blelloch:1989,Blelloch:1990}. 
At initialization of the original algorithm, these elements need to be computed so as to correspond to the following quantities:
\begin{align}
    p(\cu{x}_k \mid \cu{x}_{k-1}, y_k) 
        &= \mathcal{N}(\cu{x}_t \mid \cu{A}_k \cu{x}_{k-1} + \cu{b}_k, \cu{C}_k), \\
    p(y_k \mid \cu{x}_{k-1}) 
        &= \mathcal{N}_I(\cu{x}_{k-1} \mid \cu{\eta}_k, \cu{J}_k),
\end{align}
where $\mathcal{N}_I$ corresponds to the information form of the Gaussian distribution.
However, when no observation is available at step $k$, these equations do not hold directly and need to be modified.

By redoing the original derivation it turns out that in the case of missing measurements, the posterior $p(\cu{x}_k \mid \cu{x}_{k-1}, y_k)$ should be replaced by the transition density $p(\cu{x}_k \mid \cu{x}_{k-1}) = \mathcal{N}(\cu{x}_k \mid \cu{F}_{k-1} \cu{x}_{k-1}, \cu{Q}_{k-1})$ for $k > 1$ and $p(\cu{x}_1)$ for $k=1$. That is, the initialization equations for $\cu{A}_k, \cu{b}_k, \cu{C}_k, \cu{\eta}_k, \cu{J}_k$ in the case of missing measurements can be written as
\begin{equation}\label{eq:filter_init_k=1_missing_y}
        \cu{A}_{k} =\cu{F}_{k-1}, \quad
        \cu{b}_{k} =\cu{0}, \quad
        \cu{C}_{k} =\cu{Q}_{k-1},
\end{equation}
for $k > 1,$ and for $k = 1$ as
\begin{equation}\label{eq:filter_init_k>1_missing_y}
  \cu{A}_1 = \cu{0}, \quad
  \cu{b}_1 = \cu{0}, \quad
  \cu{C}_1 = \cu{P}_{\infty},
\end{equation}
with for all $k,$
\begin{equation}\label{eq:filter_init_information_missing_y}
        \cu{\eta}_{k} = \cu{0}, \quad
        \cu{J}_{k} = \cu{0}.
\end{equation}
%

When the quantities $\cu{A}_k, \cu{b}_k, \cu{C}_k, \cu{\eta}_k, \cu{J}_k$ have been initialized for time steps with and without observations, they can then be combined using parallel scan, with the associative operator 
defined in the same way as in \cite{Sarkka+Garcia:2021}:
\begin{align*}
\cu{A}_{ij} & =\cu{A}_{j}\left(\cu{I}_{n_{x}}+\cu{C}_{i}\cu{J}_{j}\right)^{-1}\cu{A}_{i},\\
\cu{b}_{ij} & =\cu{A}_{j}\left(\cu{I}_{n_{x}}+\cu{C}_{i}\cu{J}_{j}\right)^{-1}\left(\cu{b}_{i}+\cu{C}_{i}\cu{\eta}_{j}\right)+\cu{b}_{j},\\
\cu{C}_{ij} & =\cu{A}_{j}\left(\cu{I}_{n_{x}}+\cu{C}_{i}\cu{J}_{j}\right)^{-1}\cu{C}_{i}\cu{A}_{j}^{\top}+\cu{C}_{j},\\
\cu{\eta}_{ij} & =\cu{A}_{i}^{\top}\left(\cu{I}_{n_{x}}+\cu{J}_{j}\cu{C}_{i}\right)^{-1}\left(\cu{\eta}_{j}-\cu{J}_{j}\cu{b}_{i}\right)+\cu{\eta}_{i},\\
\cu{J}_{ij} & =\cu{A}_{i}^{\top}\left(\cu{I}_{n_{x}}+\cu{J}_{j}\cu{C}_{i}\right)^{-1}\cu{J}_{j}\cu{A}_{i}+\cu{J}_{i}.
\end{align*}
Running a parallel scan algorithm on the elements above with the filtering operator $\otimes$ then produces a sequence of ``prefix-sum'' elements $(\cu{A}^*_k, \cu{b}^*_k, \cu{C}^*_k, \cu{\eta}^*_k, \cu{J}^*_k),$ $k=1, \dots, N,$ of which the terms $\overline{\cu{x}}_{k} \triangleq \cu{b}^*_k$ and $\cu{P}_k \triangleq \cu{C}^*_k$ will then correspond to the filtering mean and covariance at time step $k$, respectively.
\begin{proposition}[Equivalence of sequential and parallel Kalman filters]
\label{prop:filter}
    For all $k=1,2,\ldots,N+M$ the Kalman filter means and covariances are given by $\overline{\cu{x}}_k = \cu{b}^*_{k}$ and $\cu{P}_k = \cu{C}^*_{k}$, respectively.
\end{proposition}
\begin{proof}
The detailed proof is omitted due to space reasons, but the result follows by explicitly writing down the forward recursion for the elements $(\cu{A}^*_k, \cu{b}^*_k, \cu{C}^*_k, \cu{\eta}^*_k, \cu{J}^*_k),$ and checking that equations for $\cu{b}^*_{k}$ and $\cu{P}_k = \cu{C}^*_{k}$ coincide with the Kalman filter equations.
\end{proof}
        
        
        On the other hand, the parallel smoothing algorithm needs no modifications compared to \cite{Sarkka+Garcia:2021} in order to handle missing observations, as it only relies on the result of the filtering algorithm and not directly on the observations.
        
\section{Temporal parallelization of Gaussian Processes}
    \label{sec:par-gps}
    
    A direct consequence of equivalence of the parallel and sequential Kalman filters and smoothers \cite{Sarkka+Garcia:2021} is the fact that the parallel and sequential versions of SSGP are equivalent too. In this section, we provide the details of the steps needed to create the linear Gaussian state-space model (LGSSM) representation of state-space GPs so that the end-to-end algorithm is parallelizable across the time dimension, and automatically differentiable, resulting a total span complexity of $O(\log N)$ in the number of observations $N$, from training to inference. 
    
    \subsection{Computation of the steady-state covariance}\label{subsec:lyapunov}
        To represent a stationary GP one must start the state-space model SDE from the stationary initial condition given by the Lyapunov equation \cite{Hartikainen+Sarkka:2010}. The complexity of this step is independent of the number of time steps and does not need explicit time-parallelization. There exist a number of iterative methods for solving this kind of algebraic equations \cite{Hodel:1996}. However, in order to make automatic differentiation efficient, in this work, we used the closed form vectorization solution given in \cite{Brogan:2011} (p. 229) which relies on matrix algebra, and does not need any explicit looping. This is feasible because we only need to numerically solve the Lyapunov equation for small state dimensions. Furthermore, as this solution only involves matrix inversions and multiplications, it is easily parallelizable on GPU architectures.
                
    \subsection{Balancing of the state space model}\label{subsec:balancing}
        In practice the state-space model obtained via discretization \eqref{eq:dmodel} is often numerically unstable due to the transition matrix having a poor conditioning number, this would in turn result in inaccuracies in computing both the GP predictions and the marginal log-likelihood of the observations. To mitigate this issue we need to resort to balancing \cite{Osborne:1960} in order to obtain a transition matrix $\cu{F}$ which has rows and columns that have equal norms, therefore obtaining a more stable surrogate state-space model. 
        For any diagonal matrix $\cu{D} \in \mathbb{R}^{n_{x} \times n_{x}},$ the continuous-discrete model
        \begin{equation}
            \begin{split}
                \frac{\dd \cu{z}(t)}{\dd t} &= \cu{D}^{-1}\,\cu{G}\, \cu{D}\, \cu{z}(t) + \cu{D}^{-1} \,\cu{L}\, \cu{w}(t),\\
                y_k &= \cu{H}\, \cu{D}\, \cu{z}_k + e_k,
            \end{split}
        \label{eq:scaled-ssmodel}
        \end{equation}
        with initial condition given by $\cu{z}(t_0) \sim \mathcal{N}(\cu{0}, \cu{D}^{-1}\,\cu{P}_\infty \cu{D})$
        is an equivalent representation of the state-space model \eqref{eq:ssmodel} with initial covariance given by $\cu{x}(t_0) \sim \mathcal{N}(\cu{0}, \cu{P}_\infty)$ with $\cu{P}_\infty$, in the sense that for all $t \ge t_0$ we have $f(t) = \cu{H}\, \cu{D}\, \cu{z}(t)$.

        In particular this means that the gradient of the log-likelihood $\log p(y_1, \ldots, y_N \mid \cu{\theta})$ with respect to the parameters $\cu{\theta}$ is left unchanged by the choice of scaling matrix $\cu{D}$. This property allows us to condition our state space representation of the original GP using the scaling matrix $\cu{D}^*$ defined in \cite{Osborne:1960}, and to compute the gradient of the log-likelihood of the observations with respect to the GP hyperparameters as if $\cu{D}^*$ did not depend on $\cu{F}$ and therefore the hyperparameters $\cu{\theta}.$ This is crucial to obtain a stable gradient while not having to unroll the gradient through the loop necessary to compute $\cu{D}^*.$ 

\subsection{Converting GPs into discrete-time state space}
    In order to use the parallel formulation of Kalman filters and smoothers in Section \ref{sec:par-kalman} we need to first form the continuous state space model representation from the initial Gaussian process definition. This operation is independent of the number of measurements and therefore has a complexity of $O(1)$. When it has been formed, we then need to transform it into a discrete-time LGSSM \eqref{eq:dmodel}. In practice, the discretization can be implemented using, for example, matrix fractions or various other methods \cite{Axelsson:2014,Sarkka+Solin:2019}. 
    These operations are fully parallelizable across the time dimension and therefore have a span complexity of $O(1)$ when run on a parallel architecture.
    
    It is worth noting that at hyperparameters learning phase, the discretization needs only to happen for the training data points. However, when predicting, it is necessary to insert the $M$ requested prediction times at the right location in the training data so as to be able to run the Kalman filter and smoother routines. When done naively, this operation has complexity $O(M+N)$ using merging operation \cite{Cormen:2009}, but can be done in parallel with span complexity $O(\log(\min(M, N))).$ In addition, for some GP models, such as Mat\'{e}rn GPs the discretization can be done offline, as it admits closed-form solutions~\cite{Sarkka+Solin:2019}.
    
\subsection{End-to-end complexity of parallelized state space GPs}
The complexity analysis of the six stages for running the parallellized state space GPs are the following:
    \begin{enumerate}
        \item Complexity of forming the continuous state-space model is $O(1)$ in both work and span complexities. 
        \item Complexity of discretizing the state-spate model is $O(N)$ is work complexity and $O(1)$ in span complexity.
        \item Complexity of the parallel Kalman filtering and smoothing operations for training is of $O(N)$ in work complexity and $O(\log N)$ is span complexity.
        \item Complexity of merging the data for the prediction is of work complexity $O(N + M)$ and of span complexity $O(\log (\min(M,N)))$. 
        \item Complexity of the parallel Kalman filtering and smoothing operations for prediction is of $O(N + M)$ in work complexity and $O(\log (N + M))$ is span complexity.
        \item Automatic differentiation as an adjoint operation has the same computational graph as the parallel Kalman filters and smoothers, and therefore has the same work and span complexities as the training step: $O(N)$ and $O(\log N)$, respectively.
    \end{enumerate}
    Putting together the above we can conclude that the total work and span complexities of doing end-to-end inference to prediction on parallelized state space GPs are $O(N + M)$ and $O(\log (N + M))$, respectively.

\section{Experiments}
\label{sec:experiments}

    In this section, we illustrate the benefits of our approach for inference and prediction on simulated and real datasets. Because GPUs are inherently massively parallelized architectures, they are not optimized for sequential data processing, exposing a lower clock speed than cost-comparable CPUs. This makes running the standard state space GPs on GPU less attractive than running them on CPU, contrarily to standard GPs which can leverage GPU-enabled linear algebra routines. In order to offer a fair comparison between our proposed methodology, the standard GPs implementation, and the standard state space GPs, we have therefore chosen to run the sequential implementation of state space GP on CPU while we run the standard GP and our proposed parallel state space GP on GPU. We verified empirically that running the standard state space GP on the same GPU architecture resulted in a tremendous performance loss for it ($\sim100$x slower), justifying running it on CPU for benchmarking. All the results were obtained using an {AMD}\textsuperscript{\textregistered} Ryzen threadripper 3960x CPU with 128 GB DDR4 RAM, and an {Nvidia}\textsuperscript{\textregistered} GeForce RTX 3090 GPU with 24GB memory.

    \subsection{Simulation model}
        We first study the behavior of our proposed methodology on a simple noisy sinusoidal model given by
        \begin{equation}\label{equ:toy-data-sin}
        \begin{split}
            f(t) &= \sin(\pi\,t) + \sin(2\,\pi\,t) + \sin(3\,\pi\,t) \\
            y_k &= f(t_k) + r_k,
        \end{split}
        \end{equation}
        with observations and prediction times being equally spaced on $(0, 4)$. By increasing the number of training or testing points we can therefore demonstrate the empirical time complexity (in terms of wall clock or work span) of our proposed method compared to the standard GP and state space GP.
        
            We have taken the kernels to be successively Mat\'{e}rn32, Mat\'{e}rn52, and RBF (approximated to the 6th order for the state space GPs), corresponding to $n_{x}=2, 3, 6$ respectively. The training dataset and test dataset have sizes ranging from $2^{12}$ to $2^{15}$. As can be seen in Figure \ref{fig:concept}, our proposed method consistently outperforms standard GP and SSGP across the chosen range of dataset sizes with standard GP eventually running out of memory for larger datasets.
    \subsection{Sunspots dataset}
        In this section we compare regression and parameter learning via likelihood maximization using L-BFGS \cite{Zhu:1997,Nocedal:1999} on the monthly sunspot activity dataset by World Data Center SILSO, Royal Observatory of Belgium, Brussels\footnote{The data is available at \url{http://www.sidc.be/silso/datafiles}}. The number of total training points in this dataset is $N = 3,200$ for learning the GP hyperparameters. We interpolate the data for every single day from 1749-01-31 to 2018-07-31 which results in 96,000 prediction points. 
        
        \begin{table}[h!]
        \caption{Running time of hyperparameter learning on sunspot dataset relative to PSSGP. The PSSGP took $39$, $46$, and $48$ milliseconds per function/gradient evaluation when $N=1200, 2200$, and $3200$, respectively.}
        \label{tbl:sunspot}
        \centering
        \begin{tabular}{@{}llll@{}}
        \toprule
        N    & GP   & SSGP  & PSSGP \\ \midrule
        1200 & 1.03 & 12.08 & $\cu{1}$     \\
        2200 & 3.82 & 25.7  & $\cu{1}$      \\
        3200 & 10.1 & 43.86 & $\cu{1}$      \\ \bottomrule
        \end{tabular}
        \end{table}
        
        The results of running time is shown in the above Table~\ref{tbl:sunspot}. PSSGP is shown to be the fastest for all $N$. Moreover, PSSGP is x10 and 43x faster than GP and SSGP when $N=3200$. Interpolating daily then took 0.14s for PSSGP, while SSGP on CPU and GP on GPU where respectively 23 and 33 times slower. 
        
\subsection{CO2 concentration dataset}
 In order to understand the impact of dimensionality of the SDE we finally consider the Mauna Loa carbon dioxide concentration dataset\footnote{The data is available at \url{https://www.esrl.noaa.gov/gmd/ccgg/trends/}} with HMC sampling \cite{Neal:2011} on the parameters. We selected monthly and weekly data from year 1974 to 2021 which contains 3192 training points. To model the periodic pattern of the data, we use the following composite covariance function $C_{\mathrm{co2}}(t-t') =  C_{\mathrm{Per.}}(t-t')\, C_\mathrm{Mat.}(t-t') + C_\mathrm{Mat.}(t-t')$ whose periodic component was converted to state space form using different orders  of its Taylor series~\cite{Solin+Sarkka:2014a}, resulting in SDEs of dimensions of $n_{x} = 10$, $14$, and $18$. This model is slightly different from the one suggested in \cite{Solin+Sarkka:2014a} where the authors also add a RBF covariance to $C_{\mathrm{co2}}$, however, we did not see any improvement in RMSE from adding this supplementary degree of freedom and therefore left it out.
 
         \begin{table}[h!]
         \caption{Relative time of sampling from the parameters posterior distribution using HMC with 10 leapfrog steps on CO2 dataset. The GP took $3$~s per sample.}
         \label{tbl:co2}
         \centering
         \begin{tabular}{@{}llll@{}}
         \toprule
         Order & GP  & SSGP  & PSSGP \\ \midrule
         1  & 1   & 4.5 & $\mathbf{0.55}$     \\
         2  & $\cu{1}$   & 5.73  & 1.36     \\
         3  & $\cu{1}$   & 6.9 & 2.55     \\ \bottomrule
         \end{tabular}
         \end{table}
        
         As we can see in Table \ref{tbl:co2}, while PSSGP is still competitive compared to SSGP for high dimensional SDE representations, its complexity increases with the dimension to the point where it eventually does not outperform the standard GP anymore.

\section{Discussion}
    We have presented a sublinear algorithm for learning and inference in state space Gaussian processes, leveraging and extending the parallel Kalman filter and smoother introduced in \cite{Sarkka+Garcia:2021}. This allowed us to reduce dramatically the training time for regression problems on large datasets as illustrated by our experiments on synthetic and real data. However, the final experiment also revealed a limitation of the methodology which is that the method does not scale well with high-dimensional state-space representations and thus more work is needed to further reduce the complexity of parallel state-space GPs. 
Recent works \cite{yaghoobi2021parallel} have suggested that this kind of parallelization techniques could also be used for non-linear state-space models, which would then make it possible to extend the present methodology for classification and deep state-space Gaussian processes~\cite{Zhao:2020}. 

\newpage
\bibliographystyle{IEEEtran}
\bibliography{main}

\end{document}


\maketitle

\section{Introduction}

The aim of this supplement is to provide a proof that the proposed parallel state-space method exactly reproduces the Gaussian process regression solution when the covariance function has a state-space representation. In the analysis, we assume that we have $N$ observations at the training points and $M$ test points without observations. We also assume that we have already sorted the observations into increasing order in time. In the Kalman filters and smoothers we then simply consider the observations at the test points as being missing (cf.\ \citep{Sarkka+Solin:2019}). The same analysis applies to both training and prediction phases as, at the training phase, we just have $M=0$ and the discretization grid is different while the filters and smoothers themselves are unchanged.

\section{Proof of Equivalence}

We start with a lemma which is a result from articles \citet{Hartikainen+Sarkka:2010,Sarkka+Solin:2013}.

\begin{lemma}[Equivalence of state-space GP and standard GP]
\label{lemma:equivalence}
Let 
\begin{equation}\label{eq:gp-def}
\begin{split}
  f(t) &\sim \mathrm{GP}(0, C(t,t')), \\
  y_{k} &= f(t_k) + e_k, \qquad e_k \sim \mathcal{N}(0, \sigma_k^2)
\end{split} 
\end{equation}
be a given Gaussian process regression problem.
If the covariance function $C(t,t')$  is taken to be stationary, in the sense that it has the form $C(t,t') = C(t-t')$ with a proper rational spectral density, then there exists a continuous-discrete linear time-invariant state-space model
\begin{equation}
\begin{split}
\frac{\dd \cu{x}(t)}{\dd t} &= \cu{G}\, \cu{x}(t) + \cu{L}\, \cu{w}(t),\\
y_k &= \cu{H} \, \cu{x}(t_k) + e_k,
\end{split}
\label{eq:ssmodel}
\end{equation}
whose smoothing distribution exactly recovers the GP regression solution at any arbitrary time point $t$. 
\end{lemma}
    
\citet{Hartikainen+Sarkka:2010,Sarkka+Solin:2013} already showed how to construct the state-space representation so that the Kalman smoother computes $p(\cu{x}(t) \mid y_{1:N}) = \mathcal{N}(\cu{x}(t) \mid \cu{m}^s(t), \cu{P}^s(t))$ such that $f(t) = \cu{H} \, \cu{x}(t) \sim \mathcal{N}(\cu{H} \, \cu{m}^s(t), \cu{H} \, \cu{P}^s(t)  \, \cu{H}^\top)$ has the same posterior mean and variance as the GP regression solution. Due to this result, given that we use the state-space model construction, it is enough to show that the parallel state-space method exactly reproduces the mean $\cu{m}^s(t)$ and covariance $\cu{P}^s(t)$ as the sequential Kalman smoother. In order to show that the parallel smoother result matches the sequential one, we need to show that the Kalman filter solutions match, because the smoother is a function of the filter result. Showing this is also important, because we have extended the method proposed in \citet{Sarkka+Garcia:2021} to account for missing observations.

Let us then recall that when there is no observation at time step $k$, the (continuous-discrete) Kalman filter (see, e.g., \citet{Sarkka+Solin:2019}) reduces to the equations
\begin{equation}
\begin{split}
  \overline{\cu{x}}_k &= \cu{F}_{k-1} \, \overline{\cu{x}}_{k-1}, \\
  \cu{P}_k &= \cu{F}_{k-1} \, \cu{P}_{k-1} \, \cu{F}_{k-1}^\top + \cu{Q}_{k-1},
\end{split}
\label{eq:nomeas}
\end{equation}
where starting from arbitrary time point $t_{k-1}$ (with or without observation) and ending to time point $t_k$ (without observation). Above, we have
\begin{equation}\label{eq:ss-gp-formation}
\begin{split}
    \cu{F}_{k-1} &= \exp\left(\left(t_k - t_{k-1}\right)\,\cu{G}\right), \\
    \cu{Q}_{k-1} &= \int_{0}^{t_k - t_{k-1}}
      \exp\left( \left(t_k - t_{k-1} - s\right)\, \cu{G}\right)
      \cu{L}\, q\,\cu{L}^\top \exp\left( \left(t_k - t_{k-1} - s\right)\,\cu{G}\right)^\top \, \dd s.
\end{split}
\end{equation}
Furthermore, when we have an observation at time step $k$ (i.e. time $t_k$), the Kalman filter equations are
\begin{equation}
\begin{split}
  \overline{\cu{x}}^-_k &= \cu{F}_{k-1} \, \overline{\cu{x}}_{k-1}, \\
  \cu{P}^-_k &= \cu{F}_{k-1} \, \cu{P}_{k-1} \, \cu{F}_{k-1}^\top + \cu{Q}_{k-1}, \\
  \overline{\cu{S}}_k &= \cu{H} \, \cu{P}_k^- \, \cu{H}^\top + \cu{R}_{k}, \\
  \overline{\cu{K}}_k &= \cu{P}_k^- \, \cu{H}^\top \overline{\cu{S}}_k^{-1}, \\
  \overline{\cu{x}}_k &= \overline{\cu{x}}_k^- + \overline{\cu{K}}_k [\cu{y}_k - \cu{H} \, \overline{\cu{x}}^-_k], \\
  \cu{P}_k &= \cu{P}_k^- - \overline{\cu{K}}_k \overline{\cu{S}}_k \overline{\cu{K}}_k^\top,
\end{split}
\label{eq:withmeas1}
\end{equation}
where we have used the matrix notation $\cu{R}_k$ for the measurement noise to retain the generalizability to multivariate observations $\cu{y}_k$. We can also write the equations as follows, which will be useful in the proofs below:
\begin{equation}
\begin{split}
  \overline{\cu{x}}_k &= \left[\cu{F}_{k-1} \,  - \cu{P}_k^- \, \cu{H}^\top \left( \cu{H} \, \cu{P}_k^- \, \cu{H}^\top + \cu{R}_{k} \right)^{-1} \cu{H} \, \cu{F}_{k-1}\right] \, \overline{\cu{x}}_{k-1} + 
  \cu{P}_k^- \, \cu{H}^\top \left( \cu{H} \, \cu{P}_k^- \, \cu{H}^\top + \cu{R}_{k} \right)^{-1} \cu{y}_k, \\
  \cu{P}_k &= \cu{P}_k^- - \cu{P}_k^- \, \cu{H}^\top \left( \cu{H} \, \cu{P}_k^- \, \cu{H}^\top + \cu{R}_{k} \right)^{-1} \, \cu{H}  \,\cu{P}_k^-.
  \end{split}
\label{eq:withmeas2}
\end{equation}

Because the parallel filtering operator is associative \cite{Sarkka+Garcia:2021}, we can select an arbitrary grouping of the operations, and the final result will be the same. We apply the operators in a sequential order starting from the beginning (step $1$) and proceeding to the end (step $N$). In that case the computation of the prefix sums $(\cu{A}^*_{k}, \cu{b}^*_{k}, \cu{C}^*_{k}, \cu{\eta}^*_{k}, \cu{J}^*_{k})$ reduces to the iteration
        \begin{align}
            &(\cu{A}^*_{k}, \cu{b}^*_{k}, \cu{C}^*_{k}, \cu{\eta}^*_{k}, \cu{J}^*_{k}) = (\cu{A}^*_{k-1}, \cu{b}^*_{k-1}, \cu{C}^*_{k-1}, \cu{\eta}^*_{k-1}, \cu{J}^*_{k-1}) \otimes (\cu{A}_{k}, \cu{b}_{k}, \cu{C}_{k}, \cu{\eta}_{k}, \cu{J}_{k}),
        \end{align} 
where $(\cu{A}_{k}, \cu{b}_{k}, \cu{C}_{k}, \cu{\eta}_{k}, \cu{J}_{k})$ were defined in Section 3.1 of the article, and the initial condition for the iteration is $(\cu{A}^*_{1}, \cu{b}^*_{1}, \cu{C}^*_{1}, \cu{\eta}^*_{1}, \cu{J}^*_{1}) = (\cu{A}_{1}, \cu{b}_{1}, \cu{C}_{1}, \cu{\eta}_{1}, \cu{J}_{1})$. We now get the following proposition.

\begin{proposition}[Equivalence of sequential and parallel Kalman filters]
\label{prop:filter}
    For all $k=1,2,\ldots,N+M$ the Kalman filter means and covariances are given by $\overline{\cu{x}}_k = \cu{b}^*_{k}$ and $\cu{P}_k = \cu{C}^*_{k}$, respectively.
\end{proposition}

\begin{proof}
At the initial step we have $\bar{\cu{x}}_1 = \cu{b}^*_1$ and $\cu{P}_1 = \cu{C}^*_1$ regardless of whether we have an observation at the step or not. Let us now assume that at step $k-1$ we have $\cu{b}^*_{k-1}= \bar{\cu{x}}_{k-1}$ and $\cu{C}^*_{k-1} = \cu{P}_{k-1}$. The recursions for $\cu{b}^*_k$ and $\cu{C}^*_k$ can then be written as
\begin{equation}
\begin{split}
          \cu{b}^*_{k}
          &= \cu{A}_{k}\left(\cu{I}_{n_{x}}+\cu{C}^*_{k-1}\cu{J}_{k}\right)^{-1}\left(\cu{b}^*_{k-1}+\cu{C}^*_{k-1}\cu{\eta}_{k}\right)+\cu{b}_{k} \\
          &= \cu{A}_{k}\left(\cu{I}_{n_{x}}+\cu{P}_{k-1}\cu{J}_{k}\right)^{-1}\left(\bar{\cu{x}}_{k-1}+\cu{P}_{k-1}\cu{\eta}_{k}\right)+\cu{b}_{k}, \\
        \cu{C}^*_{k}
        &= \cu{A}_{k} \left(\cu{I}_{n_{x}} + \cu{C}^*_{k-1}\cu{J}_{k}\right)^{-1} \cu{C}^*_{k-1} \cu{A}_{k}^{\top} + \cu{C}_{k} \\
        &= \cu{A}_{k} \left(\cu{I}_{n_{x}} + \cu{P}_{k-1}\cu{J}_{k}\right)^{-1} \cu{P}_{k-1} \cu{A}_{k}^{\top} + \cu{C}_{k}.
\end{split}
\end{equation}
If we do not have an observation at step $k > 1$, then we have $\cu{A}_{k} = \cu{F}_{k-1}$, $\cu{b}_k = \cu{0}$, $\cu{J}_k = \cu{0}$, and $\cu{\eta}_k = \cu{0}$, giving
\begin{equation}
\begin{split}
          \cu{b}^*_{k}
          &= \cu{F}_{k-1} \, \bar{\cu{x}}_{k-1}, \\
        \cu{C}^*_{k}
        &= \cu{F}_{k-1} \cu{P}_{k-1} \cu{F}_{k-1}^{\top} + \cu{Q}_{k-1},
\end{split}
\end{equation}
which indeed match Equations \eqref{eq:nomeas}. If now we do have an observation, then we have 
$\cu{A}_{k}  =\left(\cu{I}_{n_{x}}-\cu{K}_{k}\,\cu{H}\right)\cu{F}_{k-1}$, $\cu{b}_{k}  = \cu{K}_{k}\cu{y}_{k}$, and $\cu{C}_{k}  =\left(\cu{I}_{n_{x}}-\cu{K}_{k}\,\cu{H}\right)\cu{Q}_{k-1}$, where $\cu{K}_{k} =\cu{Q}_{k-1}\,\cu{H}^{\top}\,\cu{S}_{k}^{-1}$ and $\cu{S}_{k} =\cu{H}\,\cu{Q}_{k-1}\,\cu{H}^{\top}+\cu{R}_{k}$, along with $\cu{\eta}_{k} =\cu{F}_{k-1}^{\top}\,\cu{H}^{\top}\,\cu{S}_k^{-1}\,\cu{y}_k$ and $\cu{J}_{k} =\cu{F}_{k-1}^{\top}\,\cu{H}^{\top}\,\cu{S}_k^{-1}\, \cu{H}\cu{F}_{k-1}$. 

Let us first note that by matrix inversion lemma we get
\begin{equation}
\begin{split}
  \left(\cu{I}_{n_{x}} + \cu{P}_{k-1}\cu{J}_{k}\right)^{-1} &=
    \left(\cu{I}_{n_{x}} + \cu{P}_{k-1}\cu{F}_{k-1}^{\top}\,\cu{H}^{\top}\,\cu{S}_k^{-1}\, \cu{H}\cu{F}_{k-1}\right)^{-1} \\
  &= \left(\cu{I}_{n_{x}}+\cu{P}_{k-1}\cu{F}_{k-1}^{\top}\,\cu{H}^{\top}\,[\cu{H}\,\cu{Q}_{k-1}\,\cu{H}^{\top}+\cu{R}_{k}]^{-1}\, \cu{H}\cu{F}_{k-1}\right)^{-1} \\
  &= \cu{I}_{n_x} - \cu{P}_{k-1}\cu{F}_{k-1}^{\top}\,\cu{H}^{\top}\,\left(
    \cu{H}\,\cu{Q}_{k-1}\,\cu{H}^{\top}+\cu{R}_{k} + \cu{H}\cu{F}_{k-1} \cu{P}_{k-1}\cu{F}_{k-1}^{\top}\,\cu{H}^{\top} \right)^{-1} 
    \cu{H}\cu{F}_{k-1} \\
   &= \cu{I}_{n_x} - \cu{P}_{k-1}\cu{F}_{k-1}^{\top}\,\cu{H}^{\top}\,\left(
    \cu{H}\,\cu{P}^-_{k}\,\cu{H}^{\top}+\cu{R}_{k} \right)^{-1} 
    \cu{H}\cu{F}_{k-1},
\end{split}
\label{eq:invlem}
\end{equation}
where we identified $\cu{P}^-_k = \cu{F}_{k-1} \cu{P}_{k-1}\cu{F}_{k-1}^{\top} + \cu{Q}_{k-1}$ appearing in Equations~\eqref{eq:withmeas1}. We can now write
\begin{equation}
\begin{split}
  \cu{A}_k 
  &= \cu{F}_{k-1} - \cu{Q}_{k-1} \, \cu{H}^\top \, \cu{S}^{-1}_k \, \cu{H} \, \cu{F}_{k-1} \\
  &= \cu{F}_{k-1} + \cu{Q}_{k-1} \, \cu{F}_{k-1}^{-\top} \, \cu{P}^{-1}_{k-1} - \cu{Q}_{k-1} \, \cu{F}^{-\top}_{k-1} \, \cu{P}^{-1}_{k-1} (\cu{I}_{n_x} + \cu{P}_{k-1} \, \cu{F}^\top_{k-1} \, \cu{H}^\top \, \cu{S}^{-1}_k \, \cu{H} \, \cu{F}_{k-1}),
\end{split}
\label{eq:ak}
\end{equation}
where we have used the property that $\cu{F}_{k-1}$ and $\cu{P}_{k-1}$ are invertible with our model.
%
%
%
Then by using \eqref{eq:ak} and \eqref{eq:invlem} we get
\begin{equation}
\begin{split}
  \cu{A}_k \, (\cu{I}_{n_x} + \cu{P}_{k-1} \, \cu{J}_k)^{-1}
  &= (\cu{F}_{k-1} + \cu{Q}_{k-1} \, \cu{F}_{k-1}^{-\top} \, \cu{P}^{-1}_{k-1}) (\cu{I}_{n_x} + \cu{P}_{k-1} \, \cu{F}^\top_{k-1} \, \cu{H}^\top \, \cu{S}^{-1}_k \, \cu{H} \, \cu{F}_{k-1})^{-1} - \cu{Q}_{k-1} \, \cu{F}_{k-1}^{-\top} \, \cu{P}^{-1}_{k-1} \\
  &= \cu{F}_{k-1} - \cu{P}^-_k \, \cu{H}^\top \, \left(\cu{H} \, \cu{P}^-_k \, \cu{H} + \cu{R}_k \right)^{-1} \, \cu{H} \, \cu{F}_{k-1},
\end{split}
\label{eq:xcoeff}
\end{equation}
which indeed matches the coefficient of $\bar{\cu{x}}_{k-1}$ in Equations \eqref{eq:withmeas2}.
By using the matrix inversion lemma again, we further get the following:
\begin{equation}
\begin{split}
  &\cu{A}_{k}\left(\cu{I}_{n_{x}}+\cu{P}_{k-1}\cu{J}_{k}\right)^{-1} \, \cu{P}_{k-1}\cu{\eta}_{k}+\cu{b}_{k} \\
  &= \cu{A}_k (\cu{I}_{n_x} + \cu{P}_{k-1} \, \cu{J}_k)^{-1} \, \cu{P}_{k-1} \, \cu{F}_{k-1}^\top \, \cu{H}^\top \, \cu{S}^{-1}_k \, \cu{y}_k + \cu{Q}_{k-1} \, \cu{H}^\top \, \cu{S}^{-1}_k \, \cu{y}_k \\
  &= \left[ (\cu{F}_{k-1} + \cu{Q}_{k-1} \, \cu{F}_{k-1}^{-\top} \, \cu{P}^{-1}_{k-1}) (\cu{I}_{n_x} + \cu{P}_{k-1} \, \cu{F}^\top_{k-1} \, \cu{H}^\top \, \cu{S}^{-1}_k \, \cu{H} \, \cu{F}_{k-1})^{-1} - \cu{Q}_{k-1} \, \cu{F}_{k-1}^{-\top} \, \cu{P}^{-1}_{k-1} \right] \\
  &\qquad \times \cu{P}_{k-1} \, \cu{F}_{k-1}^\top \, \cu{H}^\top \, \cu{S}^{-1}_k \, \cu{y}_k + \cu{Q}_{k-1} \, \cu{H}^\top \, \cu{S}^{-1}_k \, \cu{y}_k 
  \\
  &= (\cu{F}_{k-1} + \cu{Q}_{k-1} \, \cu{F}_{k-1}^{-\top} \, \cu{P}^{-1}_{k-1}) (\cu{P}^{-1}_{k-1} + \cu{F}^\top_{k-1} \, \cu{H}^\top \, \cu{S}^{-1}_k \, \cu{H} \, \cu{F}_{k-1})^{-1} \, \cu{F}^\top \, \cu{H}^\top \, \cu{S}^{-1}_k \, \cu{y}_k \\
  &= (\cu{F}_{k-1} + \cu{Q}_{k-1} \, \cu{F}_{k-1}^{-\top} \, \cu{P}^{-1}_{k-1}) \, \cu{P}_{k-1} \, \cu{F}^\top_{k-1} \, \cu{H}^\top \, (\cu{S}_k + \cu{H} \, \cu{F}_{k-1} \, \cu{P}_{k-1} \, \cu{F}^\top_{k-1} \, \cu{H}^\top)^{-1} \, \cu{y}_k \\
  &= \cu{P}^-_k \, \cu{H}^\top \, (\cu{H} \, \cu{P}^-_k \, \cu{H}^\top + \cu{R}_k)^{-1} \, \cu{y}_k,
\end{split}
\label{eq:ycoeff}
\end{equation}
which has the same coefficient for the observation as in Equations \eqref{eq:withmeas2}. From Equations \eqref{eq:xcoeff} and \eqref{eq:ycoeff} we can thus conclude that $\overline{\cu{x}}_k = \cu{b}^*_k$.

Similarly, by using matrix inversion lemma and some simplifications, we arrive at
\begin{equation}
\begin{split}
\cu{A}_{k} \left(\cu{I}_{n_{x}} + \cu{P}_{k-1}\cu{J}_{k}\right)^{-1} \cu{P}_{k-1} \cu{A}_{k}^{\top} + \cu{C}_{k}
&= \cu{P}^-_k - \cu{P}^-_k \, \cu{H}^\top \, (\cu{H} \, \cu{P}^-_k \, \cu{H} + \cu{R}_k)^{-1} \ \cu{H} \, \cu{P}^-_k,
\end{split}
\end{equation}
which shows that $\cu{C}^*_{k} = \cu{P}_k$. The result then follows by an induction argument.
\end{proof}

We can now turn our attention to the smoother. The missing observations do not affect the form of the smoothing equations (cf.\ \citet{Sarkka+Solin:2019}) and thus it is enough to consider the standard smoother which has the form
\begin{equation}
\begin{split}
  \cu{G}_k &= \cu{P}_k \, \cu{F}_k^\top \, \left( \cu{F}_k \cu{P}_k \cu{F}_k^\top + \cu{Q}_k \right)^{-1}, \\
  \cu{m}^s_k &= \overline{\cu{x}}_k + \cu{G}_k (\cu{m}^s_{k+1} - \cu{F}_k \overline{\cu{x}}_k), \\
  \cu{P}^s_k &= \cu{P}_k - \cu{G}_k (\cu{P}^s_{k+1} - \cu{F}_k \cu{P}_k \cu{F}_k^\top - \cu{Q}_k) \cu{G}_k^\top.
\end{split}
\label{eq:smoother}
\end{equation}
Again, because the smoothing operator is associative, we can choose any grouping of the operations, and here we elect to use a backward iteration:
        \begin{align}
            (\cu{E}^*_{k}, \cu{g}^*_{k}, \cu{L}^*_{k}) = (\cu{E}_{k}, \cu{g}_{k}, \cu{L}_{k}) \otimes (\cu{E}^*_{k+1}, \cu{g}^*_{k+1}, \cu{L}^*_{k+1})
        \end{align}
which is started from $(\cu{E}^*_{N+M}, \cu{g}^*_{N+M}, \cu{L}^*_{N+M}) = (\cu{E}_{N+M}, \cu{g}_{N+M}, \cu{L}_{N+M})$.

\begin{proposition}[Equivalence of sequential and parallel Kalman smoothers]
\label{prop:smoother}
    For all $k=1,2,\ldots,N+M$ the Kalman smoother means and covariances are given by $\cu{m}^s_k = \cu{g}^*_{k}$ and $\cu{P}^s_k = \cu{L}^*_{k}$ respectively.
\end{proposition}

\begin{proof}
At the final step $k=N+M$ this indeed is true. Then if we assume that it is true for step $k+1$, we get
\begin{equation}
\begin{split}
            \cu{g}^*_{k} & =\cu{E}_{k}\cu{g}_{k+1}+\cu{g}_{k}\\
            &= \cu{E}_{k} m^s_{k+1} +\cu{g}_{k}, \\
            \cu{L}^*_{k} & =\cu{E}_{k} \cu{L}_{k+1}\cu{E}_{k}^{\top}+\cu{L}_{k} \\
            & =\cu{E}_{k} \cu{P}^s_{k+1}\cu{E}_{k}^{\top}+\cu{L}_{k}.
\end{split}
\end{equation}
We can also notice that $E_k = G_k$ is exactly the smoother gain and thus we get
\begin{equation}
\begin{split}
            \cu{g}^*_{k} 
            &= \cu{E}_{k} \cu{m}^s_{k+1} +\cu{g}_{k} \\
            &= \cu{G}_{k} \cu{m}^s_{k+1} + \overline{\cu{x}}_{k}-\cu{G}_{k}\left(\cu{F}_{k}\,\overline{\cu{x}}_{k}\right) \\
            &= \overline{\cu{x}}_{k} + \cu{G}_{k} (\cu{m}^s_{k+1} - \cu{F}_{k}\,\overline{\cu{x}}_{k})
\end{split}
\end{equation}
and
\begin{equation}
\begin{split}
            \cu{L}^*_{k} 
            &= \cu{E}_{k} \cu{P}^s_{k+1}\cu{E}_{k}^{\top}+\cu{L}_{k} \\
            &= \cu{G}_{k} \cu{P}^s_{k+1}\cu{G}_{k}^{\top}+\cu{P}_{k}-\cu{G}_{k}\,\cu{F}_{k}\,\cu{P}_{k} \\
            &= \cu{G}_{k} \cu{P}^s_{k+1}\cu{G}_{k}^{\top}+\cu{P}_{k}-\cu{G}_{k}\,(\cu{F}_{k}\,\cu{P}_{k}\cu{F}_{k}^\top + \cu{Q}_k) \cu{G}_{k}^\top \\
            &= \cu{P}_{k} + \cu{G}_{k} (\cu{P}^s_{k+1} - \cu{F}_{k}\,\cu{P}_{k}\cu{F}_{k}^\top - \cu{Q}_k) \cu{G}_{k}^\top,
\end{split}
\end{equation}
which are precisely the smoother equations \eqref{eq:smoother}, and hence $\cu{g}^*_{k} = \cu{m}^s_k$ and $\cu{L}^*_{k} = \cu{P}^s_k$. The result follows by an induction argument.
\end{proof}

We now get the following corollary by combining Lemma~\ref{lemma:equivalence} and Propositions~\ref{prop:filter} and \ref{prop:smoother}.

\begin{corollary}
As a consequence the parallel and sequential state space representations of GPs are equivalent. Therefore, for covariance functions such that the sequential state-space GP is an exact representation of the standard GP (in the sense of Lemma~\ref{lemma:equivalence}), the parallel state-space GP will be an exact representation as well. 
\end{corollary} 

\bibliography{refs}